\documentclass{article}

\usepackage{arxiv}

\usepackage[utf8]{inputenc} 
\usepackage[T1]{fontenc}    
\usepackage{hyperref}       
\usepackage{url}            
\usepackage{booktabs}       
\usepackage{amsfonts}       
\usepackage{nicefrac}       
\usepackage{microtype}      
\usepackage{lipsum}		
\usepackage{graphicx}
\usepackage{natbib}
\usepackage{doi}
\usepackage{xcolor, soul}

\title{IAMAP: Unlocking Deep Learning in QGIS for non-coders and limited computing resources}

\author{ 
\href{https://orcid.org/0000-0002-1275-4673}{\includegraphics[scale=0.06]{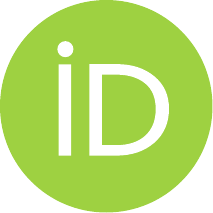}}\hspace{1mm}%
Paul~Tresson$^{*,1}$,
Pierre~Le~Coz$^{1,2}$,%
Hadrien~Tulet$^{1}$,%
\href{https://orcid.org/0000-0001-9603-4448}{\includegraphics[scale=0.06]{orcid.pdf}}\hspace{1mm}%
Anthony~Malkassian$^{3}$,%
\href{https://orcid.org/0000-0003-2824-267X}{\includegraphics[scale=0.06]{orcid.pdf}}\hspace{1mm}%
Maxime~Réjou-Méchain$^{1,2}$%
\vspace{0.5em}
\\
$^1$ AMAP, Univ. Montpellier, IRD, CNRS, CIRAD, INRAE, Montpellier, France\\
$^2$ Forest Restoration Research Unit, Department of Biology, Faculty of Science,\\ Chiang Mai University, Chiang Mai, Thailand\\
$^3$Université de la Réunion, UMR PVBMT, St. Pierre, La Réunion, France\\
\vspace{0.5em}
}

\begin{document}
\maketitle

\begin{abstract}

\textbf{1.} Remote sensing has entered a new era with the rapid development of artificial intelligence approaches. However, the implementation of deep learning has largely remained restricted to specialists and has been impractical because it often requires (i) large reference datasets for model training and validation; (ii) substantial computing resources; and (iii) strong coding skills.

\textbf{2.} Here, we introduce IAMAP, a user-friendly QGIS plugin that addresses these three challenges in an easy yet flexible way. IAMAP builds on recent advancements in self-supervised learning strategies, which now provide robust feature extractors, often referred to as foundation models. These generalist models can often be reliably used in few-shot or zero-shot scenarios (\textit{i.e.}, with little to no fine-tuning).

\textbf{3.} IAMAP’s interface allows users to streamline several key steps in remote sensing image analysis: (i) extracting image features using a wide range of deep learning architectures; (ii) reducing dimensionality with built-in algorithms; (iii) performing clustering on features or their reduced representations; (iv) generating feature similarity maps; and (v) calibrating and validating supervised machine learning models for prediction.

\textbf{4.} By enabling non-AI specialists to leverage the high-quality features provided by recent deep learning approaches without requiring GPU capacity or extensive reference datasets, IAMAP contributes to the democratization of computationally efficient and energy-conscious deep learning methods.

\end{abstract}

\newpage

\section{Introduction}

The integration of remote sensing data with deep learning approaches is currently revolutionizing Earth observation sciences, leading to significant qualitative and quantitative improvements in large-scale predictions \citep{zhu2017deep,yuan2020deep,yasir2023coupling}. However, this revolution comes with a number of challenges. First, over the past decade, most deep learning applications have been highly data-demanding, requiring extensive manual labeling with typically more than one hundred thousands labeled points \citep{safonova2023ten}. In most ecological and environmental science studies, constructing such a large reference dataset, through \textit{e.g.}, ground observations or photo-interpretation, remains a major barrier to the implementation of deep learning approaches. Second, a common obstacle to the adoption of deep learning is the computing power required to train a model. Training a deep learning model is indeed highly resource-intensive, primarily due to the backpropagation step \citep[see][]{goodfellow2016deep}. As a result, modern deep learning architectures are virtually impossible to train without substantial local Graphics Processing Unit (GPU) capacity or access to high-end computing clusters. Last but not least, implementing deep learning approaches typically requires at least basic coding skills, which has so far restricted their use to users with a minimal background in computer science.

The recent development of self-supervised learning (SSL) approaches is a game-changer in the deep learning domain, as exemplified by the success of models like BERT and ChatGPT in natural language processing \citep{devlin2018bert, achiam2023gpt}. 
In SSL, the model starts by learning features describing a dataset via a pretext task that does not require a label. 
In computer vision, several SSL strategies have been proposed, typically belonging to two main categories: contrastive or generative learning. In contrastive learning, several networks view transformed versions of the same data and have to learn to produce robust representation of this data (\textit{e.g.} DINOv2 \cite{oquab2023dinov2} or VicReg \cite{bardes2021vicreg}). In generative learning, a network sees a degraded version of the data (typically, a masked version) and has to learn to generate a non-degraded version (\textit{e.g.}, MAE, \cite{he2022masked}) \citep[for an overview of main SSL approaches, see ][]{shwartz2024compress}. 
Once pre-trained on a large set of images, which remains very data- and resource-intensive, the resulting backbone can be referred to as a "foundation model". Like any pre-trained model, this foundation model can then be fine-tuned with a limited number of manually labeled examples to learn a specific downstream task (\textit{e.g.} land cover classification or change detection in remote sensing) \citep{ericsson2021well}. 
The main difference between a pre-trained self-supervised learning (SSL) model and a pre-trained supervised model lies in their training objectives: SSL models are not constrained by predefined labels and are therefore free to explore and encode the intrinsic structure and diversity of the data, often resulting in more general and transferable representations. In contrast, supervised models are explicitly optimized to perform a specific user-defined task, which can lead to highly specialized representations that may overlook other meaningful features in the data.
As such, SSL foundation models can perform well even in low-shot or zero-shot tasks, \textit{i.e.} using the model as is, with few or no training data. Consequently, SSL models are considered particularly promising for remote sensing tasks, as demonstrated by recent works and initiatives \citep{jakubik2023prithvi, cong2023satmae, xiong2024neural, marsocci2024pangaea}.

In parallel, to the development of SSL, Vision Transformers (ViT) \citep{dosovitskiy2020image}  and their derivatives (such as EVA \citep{fang2023eva} or Hiera \citep{ryali2023hiera}) have changed the state of the art of  computer vision. In a ViT, an image is analyzed by patches (usually $16\times16$ or $14\times14$ pixels). Each patch is projected in an embedding space and the embedding spaces of the different patches update each-others through the self-attention mechanism \citep[see ][]{vaswani2017attention}.  These architectures present the advantage that the features produced at patch level can be analyzed spatially within an image, which is relevant in remote sensing especially when working at high resolution, (see \cite{marsocci2024pangaea}) (see \href{https://iamap.readthedocs.io/en/latest/faq.html\#how-does-a-vit-work}{the plugin documentation} for a more detailed overview of the functioning of a ViT).

With the democratization of deep learning, some developers have already worked on the integration of deep learning models in geographic information systems such as the open-source and widely used QGIS software \cite{QGIS}.
However, at the time of writing, these solutions mostly focus on fine-tuning models or using a model in inference only \citep[\textit{e.g.} see ][]{aszkowski2023deepness, zhao2023geosam}. Then, they are only usable by users with access to high-end computing power, extensive dataset, on interested in a task for which a specific model was already trained.

In this paper, we introduce a new plugin for QGIS designed to streamline remote sensing image analysis using advanced pre-trained deep learning models without the need for coding or extensive computing resources. As demonstrated in this paper, users can apply a pre-trained model to generate high-quality features at the patch level. The plugin then allows users to manipulate these features using various projections, clustering, similarity, and supervised machine learning (ML) algorithms.

\section{Plugin description}

\begin{figure}
    \centering
    \includegraphics[width=0.9\linewidth]{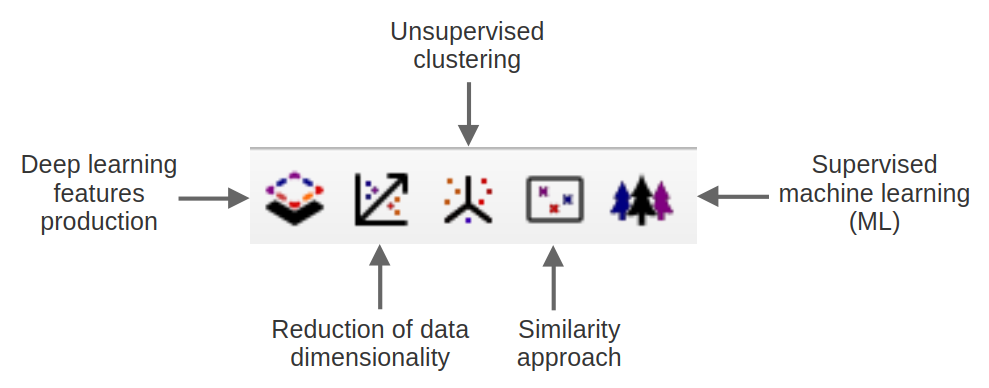}
    \caption{The five main modules of the IAMAP plugin.}
    \label{fig:functionalities}
\end{figure}

The IAMAP plugin integrated into QGIS consists of five main modules, which can be used individually or sequentially on a georeferenced raster image (Fig.~\ref{fig:functionalities}). We here below describe the functionality of each module.

\subsection{Deep Learning feature production}

The first and most original module of IAMAP is the deep learning feature extraction module. Using a georeferenced raster as input (a QGIS raster layer or a raster saved on disk), this module enables the use of various pre-trained deep learning models to produce a set of features describing the input raster.  The use of deep learning model in inference only removes the costly training step and greatly reduce the computational power required.
This module mostly relies on two widely used \textit{PyTorch} libraries: \textit{timm} \citep{rw2019timm}, for loading pre-trained model weights, and \textit{torchgeo} \citep{torchgeo2022}, for handling geospatial data. 

The \textit{timm} library has become a standard for sharing and loading pre-trained weights in \textit{PyTorch} and is now integrated into the HuggingFace Hub (\url{https://huggingface.co/},\cite{wolf2019huggingface}). Originally developed for sharing natural language processing (NLP) models, the HuggingFace Hub has since become the largest repository of pre-trained deep learning models, with over 400,000 models available at the time of writing. Our aim in choosing this back-end is to rely on libraries that are widely used, well maintained, and actively updated. Hence, while we propose a couple of widely used foundation models by default, the user can select any model available on HuggingFace by entering the architecture name (although not all models are guaranteed to work depending on their architecture). The plugin interface also gives the possibility to load local pre-trained models weights, if a correct \textit{timm} architecture is chosen.

Handling remote sensing datasets differs from working with typical image collections used in classical computer vision. Raster images are often several orders of magnitude larger and must be sampled to fit the input requirements of neural networks, which typically expect square images a few hundred pixels wide. Additionally, it is essential to preserve the geographical metadata associated with each raster. The \textit{torchgeo} library provides an efficient solution to address these constraints, but it includes many features and dependencies that are not all useful for our purposes. Therefore, we have forked only the necessary parts of the \textit{torchgeo} code into our plugin. Our goal is to keep the codebase simple and minimize unnecessary dependencies.

The module offers several options to the user, most of which come with proposed default values. Among these, the sampling size and the stride are key parameters: the sampling size determines the dimensions of the extracted tiles while the stride controls the spacing between tiles and thus the degree of overlap used to reduce tiling artifacts. The combination of sampling size, stride and the architecture chosen as encoder will determine the resolution of the output raster. It is possible as well to set an overlap between tiles to reduce possible tilling effects. These parameters are essential to consider, as they directly influence the trade-off between model performance and inference cost.

\begin{figure}
    \centering
    \rotatebox{90}{Original image}
    \includegraphics[width=0.4\linewidth]{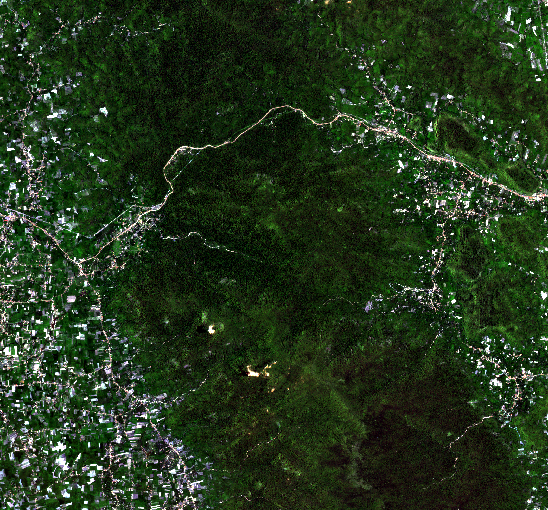}
    \vspace{.5cm}
    
\begin{tabular}{lccc}

     &
     ViT base DINO&
     ViT base MAE&
     SSL4EO DINO\\
     
     &
     \small{\cite{caron2021emerging}}&
     \small{\cite{he2022masked}}&
     \small{\cite{wang2022ssl4eo}}\\
     
     \rotatebox{90}{Raw features} & 
     \includegraphics[width=0.3\linewidth]{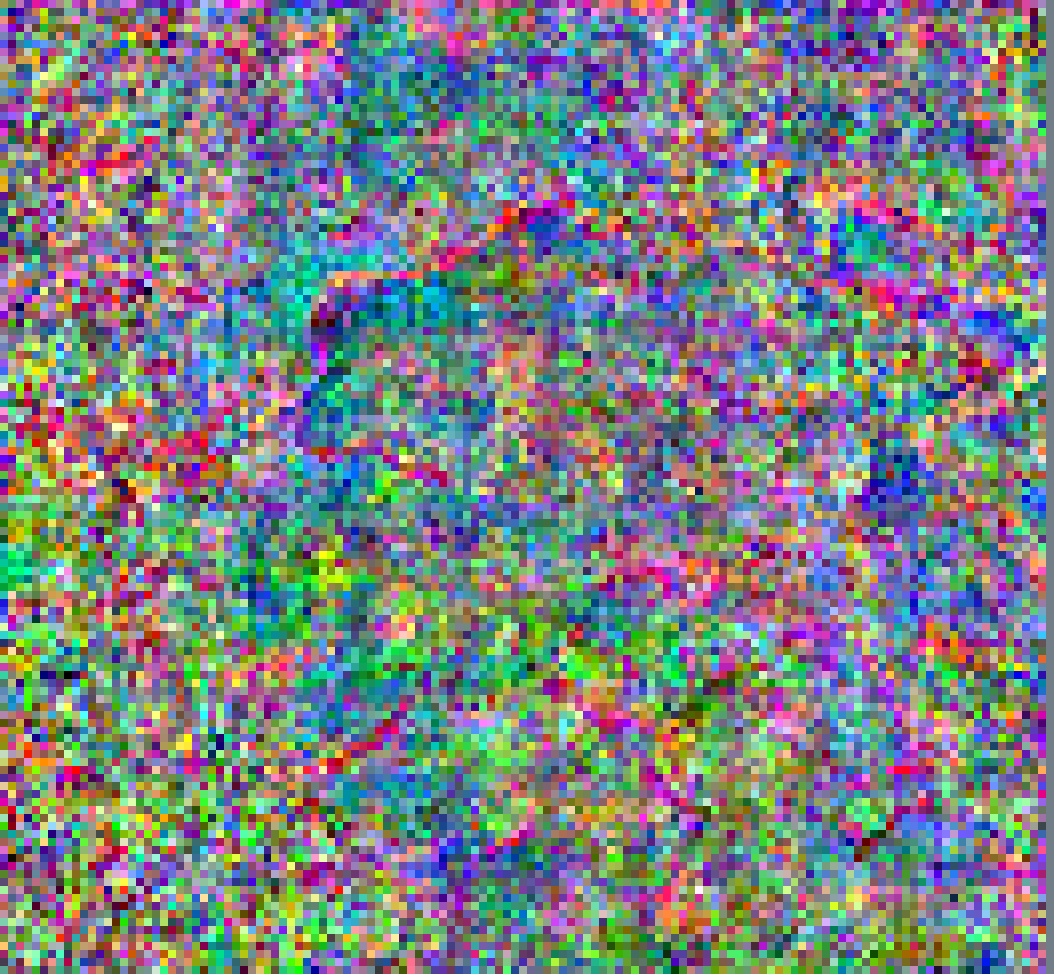} &
     \includegraphics[width=0.3\linewidth]{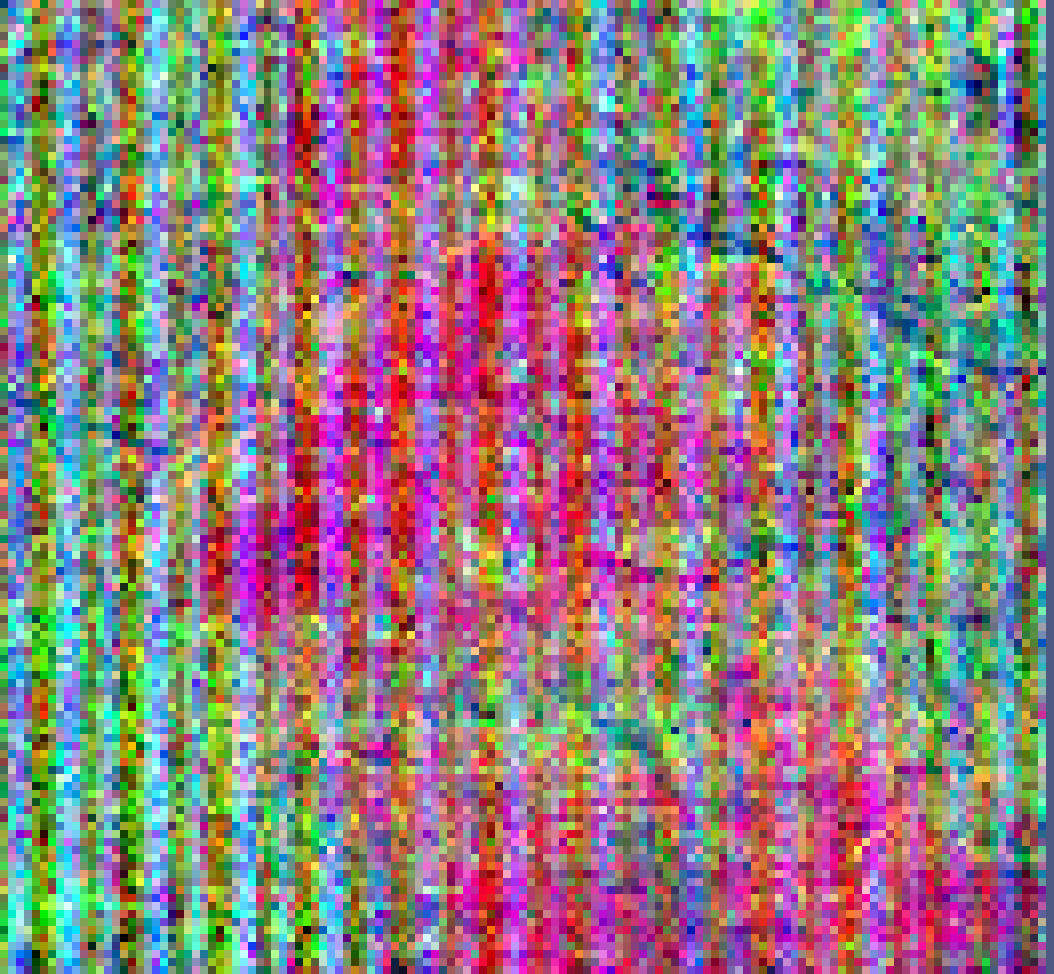} &
     \includegraphics[width=0.3\linewidth]{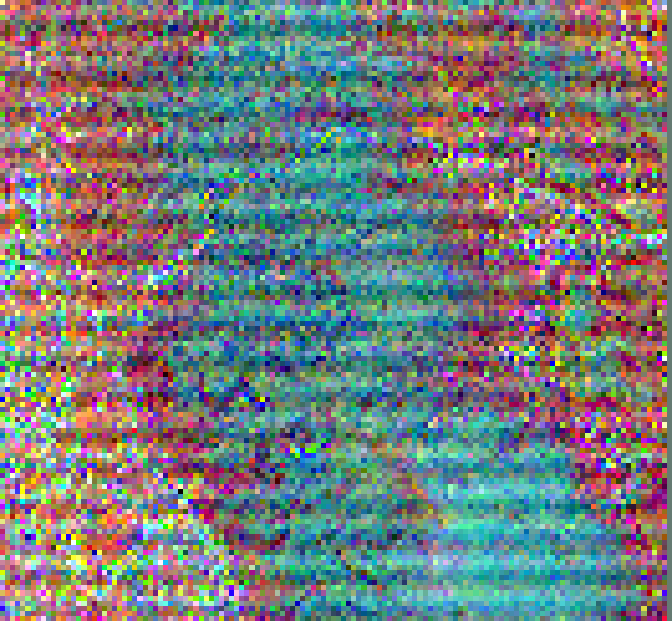}\\

     \rotatebox{90}{PCA} &
     \includegraphics[width=0.3\linewidth]{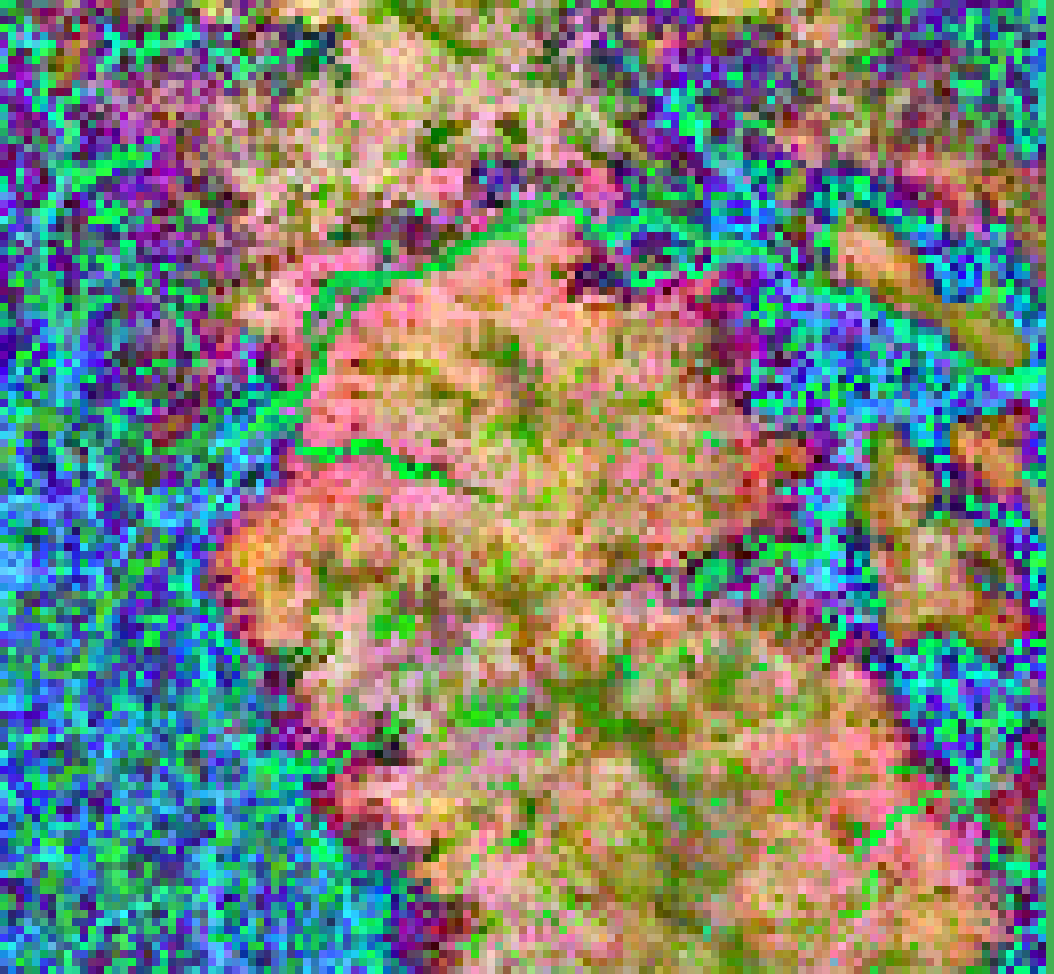} &
     \includegraphics[width=0.3\linewidth]{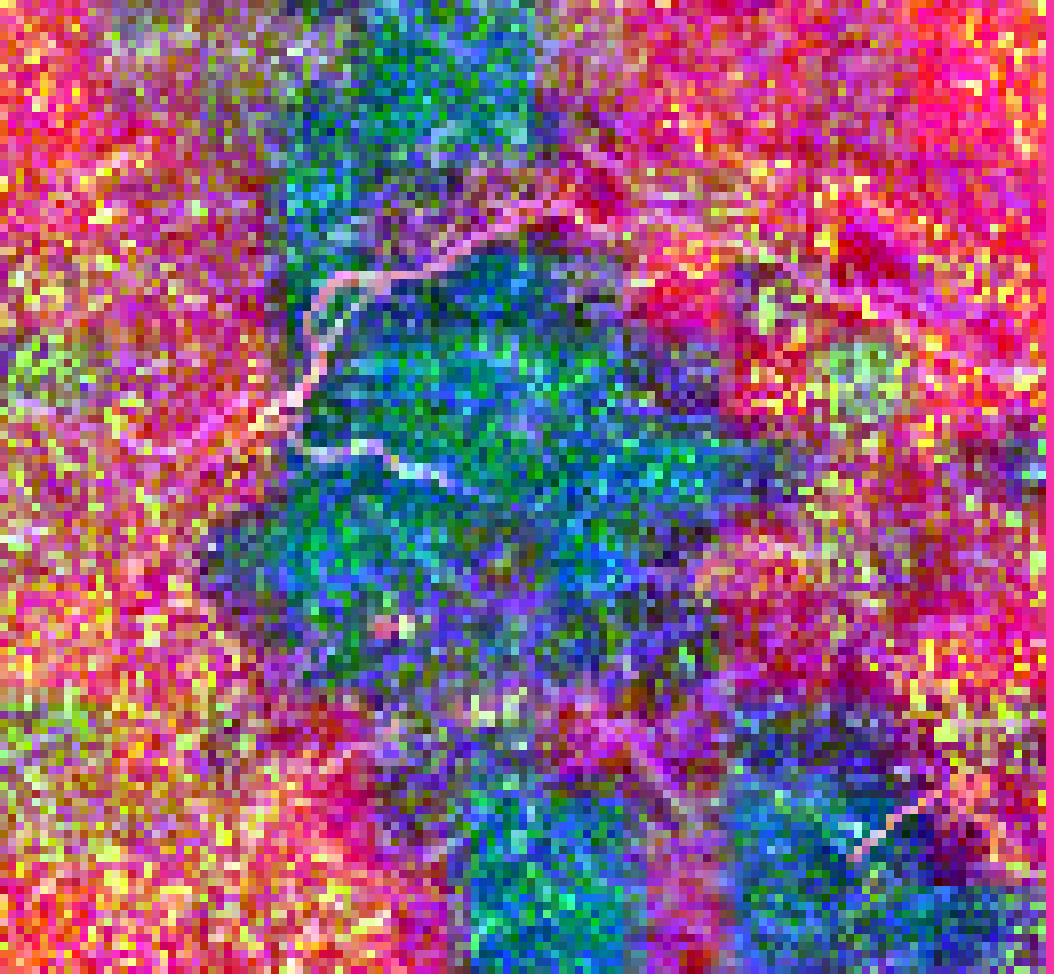} &
     \includegraphics[width=0.3\linewidth]{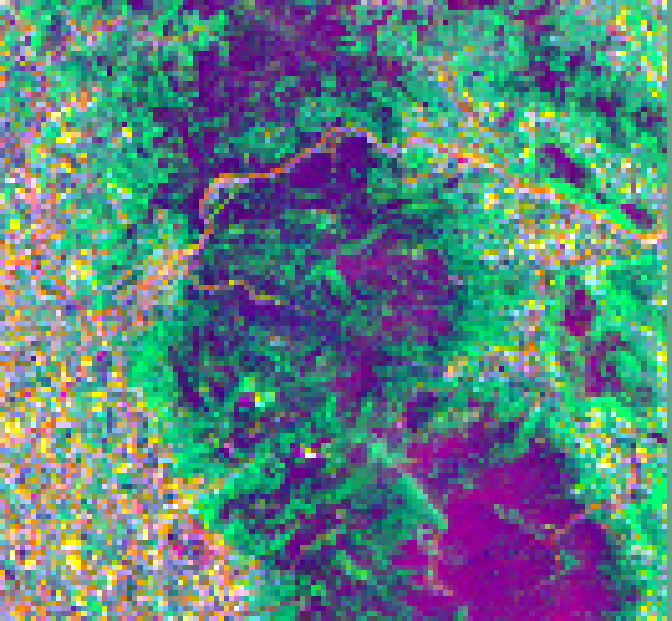}\\

     \rotatebox{90}{T-SNE} &
     \includegraphics[width=0.3\linewidth]{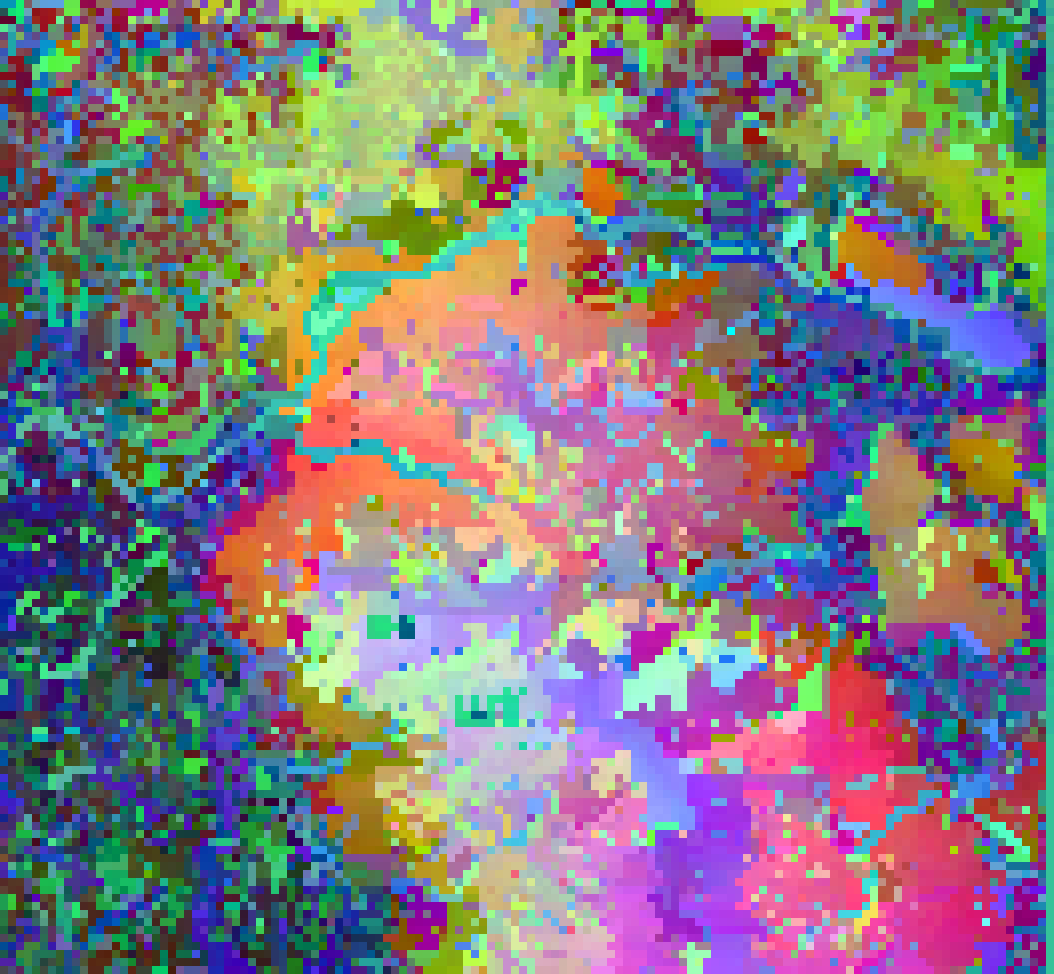} &
     \includegraphics[width=0.3\linewidth]{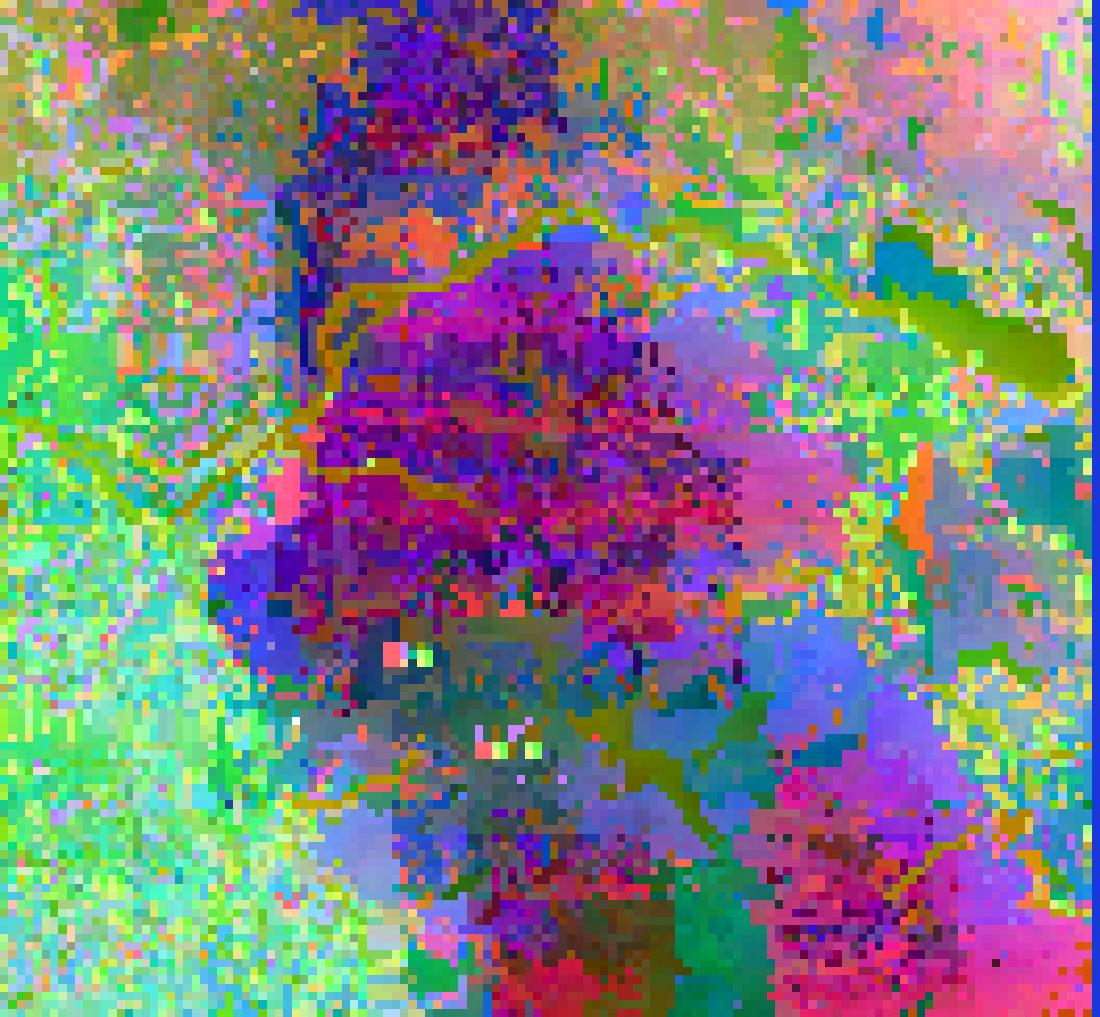} &
     \includegraphics[width=0.3\linewidth]{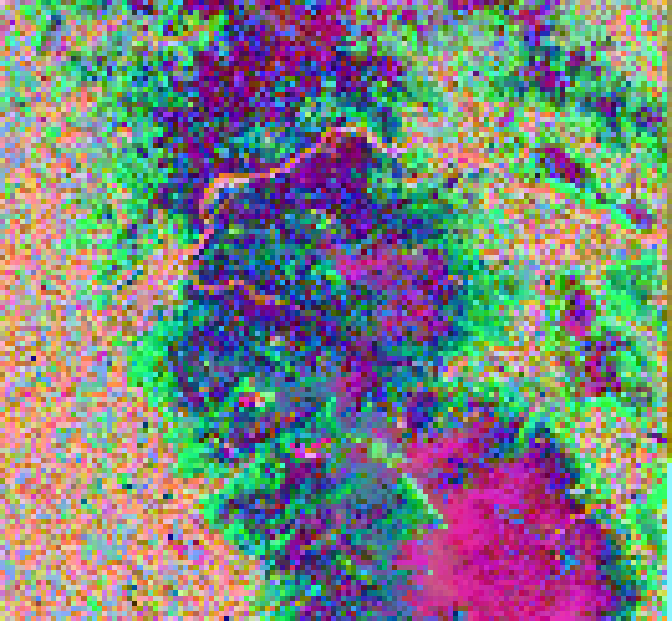}\\
\end{tabular}
    \caption{A sentinel 2 image of a forested landscape in Thailand (Khao Banthat Wildlife Sanctuary; Lat 7.53°, Lon 99.82°) processed by different backbones. The top row represents the first three feature dimensions output by the models (which may not be the most informative). The second row shows a 3D PCA of the features mapped to the red, green and blue channel respectively. The third row shows a projection using a 3D T-SNE.}.
    \label{fig:backbones}
\end{figure}

The output of this module is a raster with a coarser resolution than the input raster, depending on the sampling parameters and the chosen deep learning architecture. It contains as many bands as the number of extracted features (\textit{e.g.} 768 for a ViT-base model). By default, QGIS loads the raster at the end of the process and displays only the first three bands using a false-color RGB composition, although these bands are not necessarily the most informative (see top row of Fig. \ref{fig:backbones}).

Computer vision state of the art pretrained models are usually trained with Red Green and Blue (RGB) bands used in natural images. We thus propose three strategies for users who want to work with input rasters with a band number different from 3, as it is usually the case in remote sensing. One potential solution requires manipulations of the pre-trained weights to handle the exact number of bands of the raw image by either copying the weights of the first layer modulo 3 if the number of input bands is larger than three or averaging weights if the number of input bands is smaller than three. This option should be taken with caution given that it is expected to change the behavior of the model, even if it should keep a capacity for abstraction and projecting low level information into a richer feature space (see Figure \ref{fig:backbones} examples). The second solution consists in selecting only 3 relevant bands in the deep learning module without modifying the model's weights. The last option, which appears to be the most robust one according to our tests, consists in applying first a dimension reduction (see next section), such as a PCA, and use three reduction axes as an input in the deep learning encoder.

As the state of the art is evolving for remote sensing application, we also provide inference with foundation models trained specifically on remote sensing data, such as DOFA \citep{xiong2024neural} and a ViT pretrained on the SSL4EO dataset \citep{wang2022ssl4eo} using \citeauthor{marsocci2024pangaea} implementation \citep{marsocci2024pangaea}. For the DOFA encoder, multispectral bands are handled by the model without manipulation of the encoder or pre-processing required.

\subsection{Reduction of data dimensionality}

This module enables dimensionality reduction of an input raster using a variety of approaches, including PCA, t-SNE \citep{van2008visualizing}, and UMAP \citep{mcinnes2018umap}. This dimensionality reduction step is particularly useful for two tasks: (1) reducing the number of bands in a raw multi-band raster before applying a deep learning model, as discussed in the previous section, and (2) reducing the dimensionality of the feature space to facilitate visualization and support more robust training procedures. Indeed, deep learning models typically produce a high-dimensional feature space. While this high dimensionality poses no issues when fed into a deep learning head, it can become a drawback for visualizing the feature space and using it in lighter machine learning models such as Random Forests. To address this, it is common in deep learning research to use dimensionality reduction algorithms to visualize and analyze the feature space of a model. These reduced features can often be more informative at first glance (see the second row of Fig.~\ref{fig:backbones}), and reducing or ordering the input dimensions can improve the performance of other algorithms afterward (see the third row of Fig.~\ref{fig:backbones}).

This module relies on the \textit{scikit-learn} library, which provides access to a wide range of algorithms (25 at the time of writing). As a result, all algorithms available in the \textit{scikit-learn} \textit{decomposition} and \textit{cluster} modules that have common APIs (namely, a \textit{fit()}, a \textit{transform()}, or a \textit{fit\_transform()} method) can be used.  Note that the UMAP approach relies instead on its dedicated Python implementation and is an optional dependency at the time of writing.

\subsection{Unsupervised clustering}

\begin{figure}
    \centering
\begin{tabular}{ccc}
     K-means&
     K-Means&
     Spectral Clustering\\
     &
     after 3D T-SNE&
     after 3D T-SNE\\
     
     \includegraphics[width=0.3\linewidth]{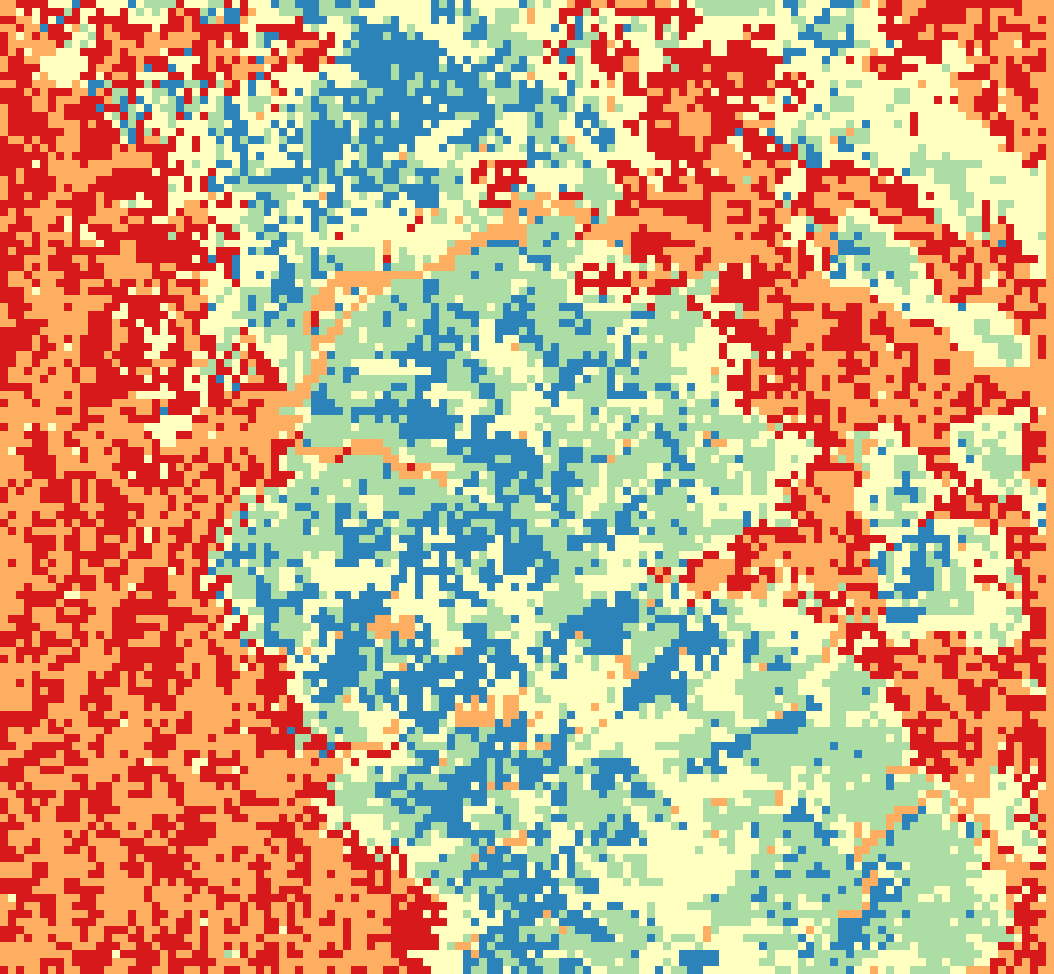} &
     \includegraphics[width=0.3\linewidth]{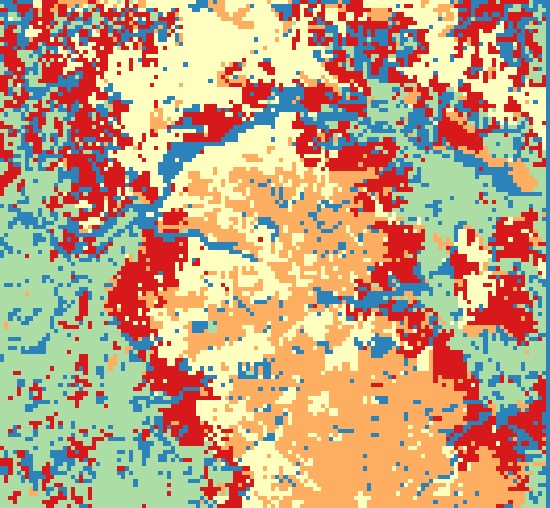} &
     \includegraphics[width=0.3\linewidth]{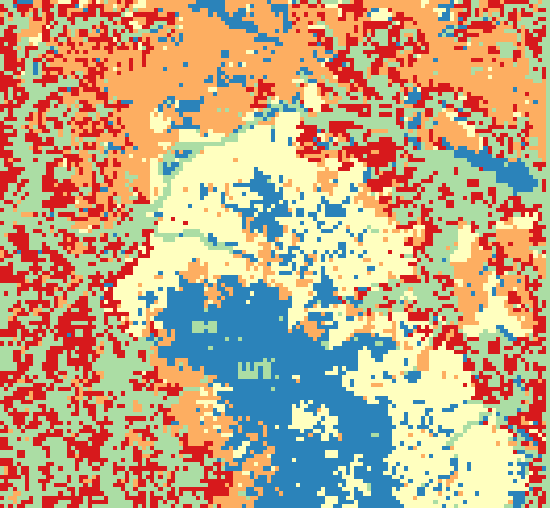}\\
\end{tabular}
    \caption{Example of different clustering (k=5) of the ViT Base DINO features.}
    \label{fig:clusterings}
\end{figure}

A common operation when handling feature spaces is clustering to assign classes to data points. The unsupervised clustering module allows to implement various unsupervised clustering algorithms, including K-means or HDBSCAN \citep{mcinnes2017hdbscan} (see Fig.~\ref{fig:clusterings}). This module again relies on \textit{scikit-learn} as a back-end. As such, all algorithms available in the \textit{scikit-learn} \textit{cluster} module sharing common APIs (namely, a \textit{fit()}, a \textit{predict()}, or a \textit{fit\_predict()} method) can be used.

\subsection{Similarity approach}

\begin{figure}
    \centering
\begin{tabular}{ccc}
     
     \includegraphics[width=0.45\linewidth]{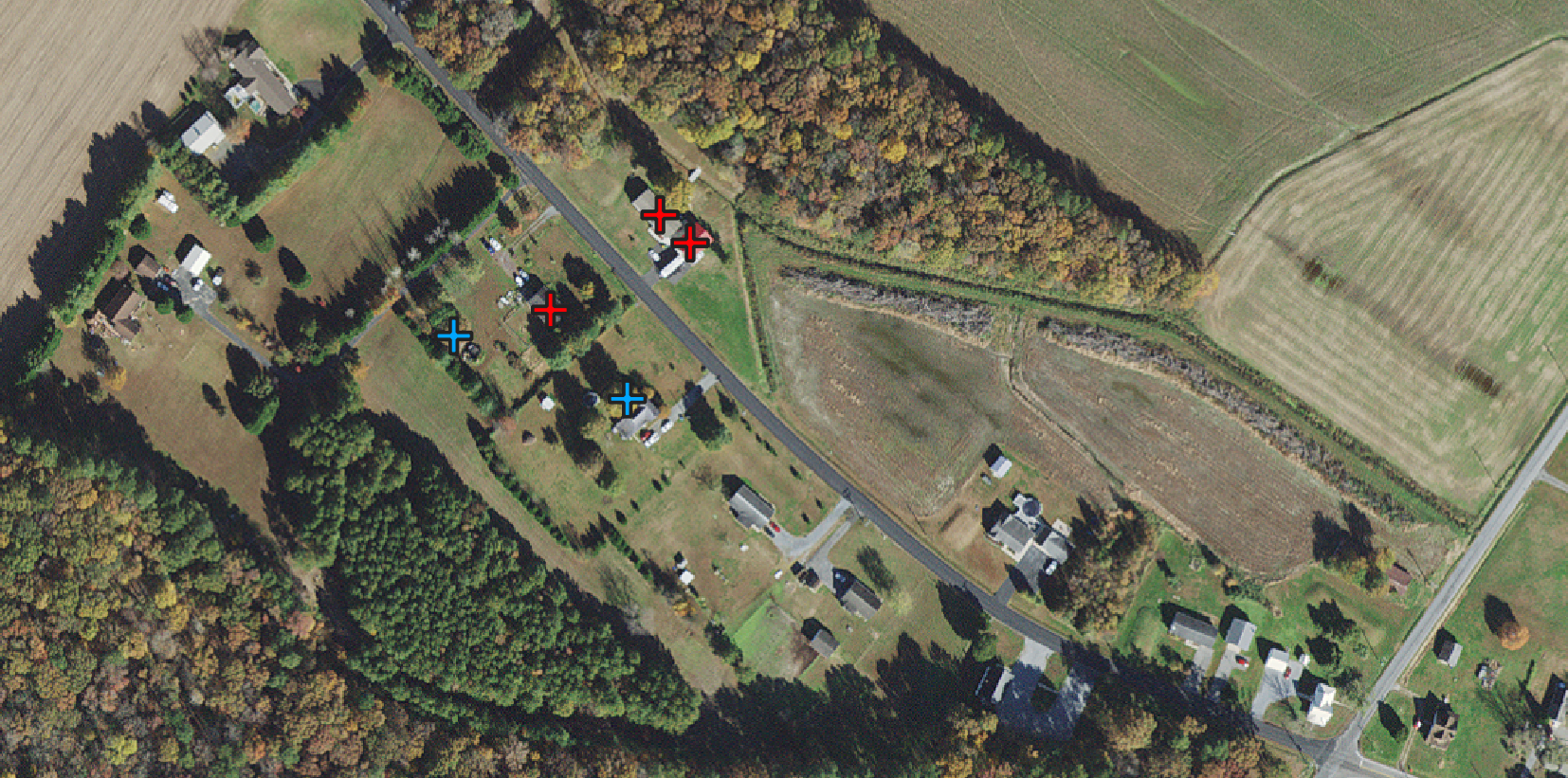} & &
     \includegraphics[width=0.45\linewidth]{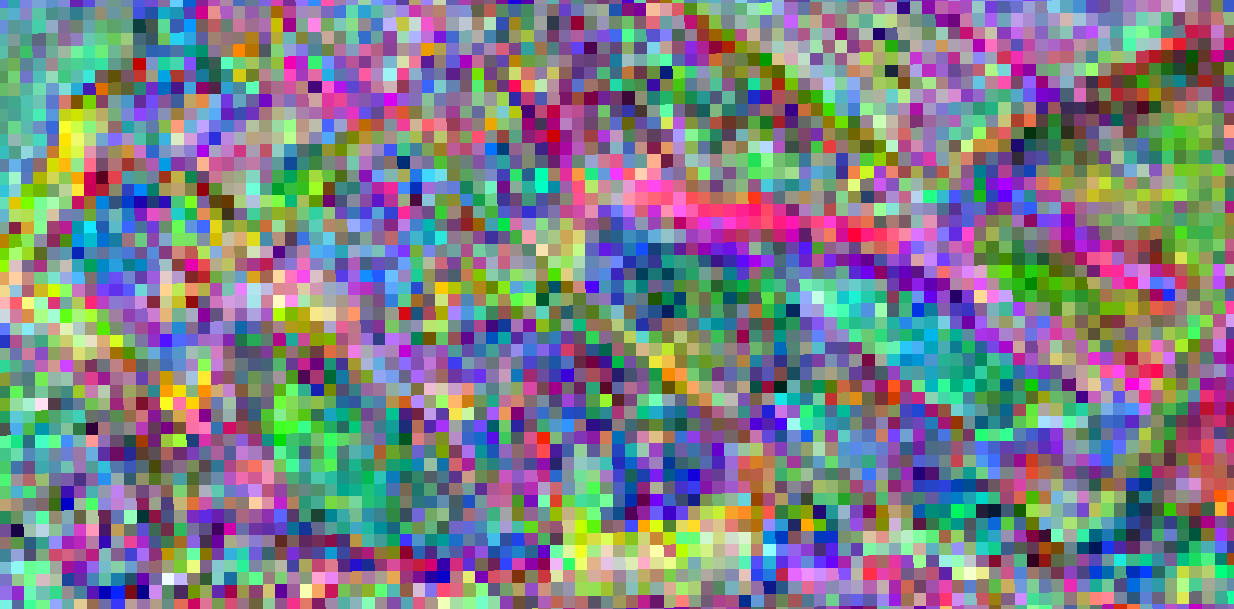} \\
     & & \\
     \includegraphics[width=0.45\linewidth]{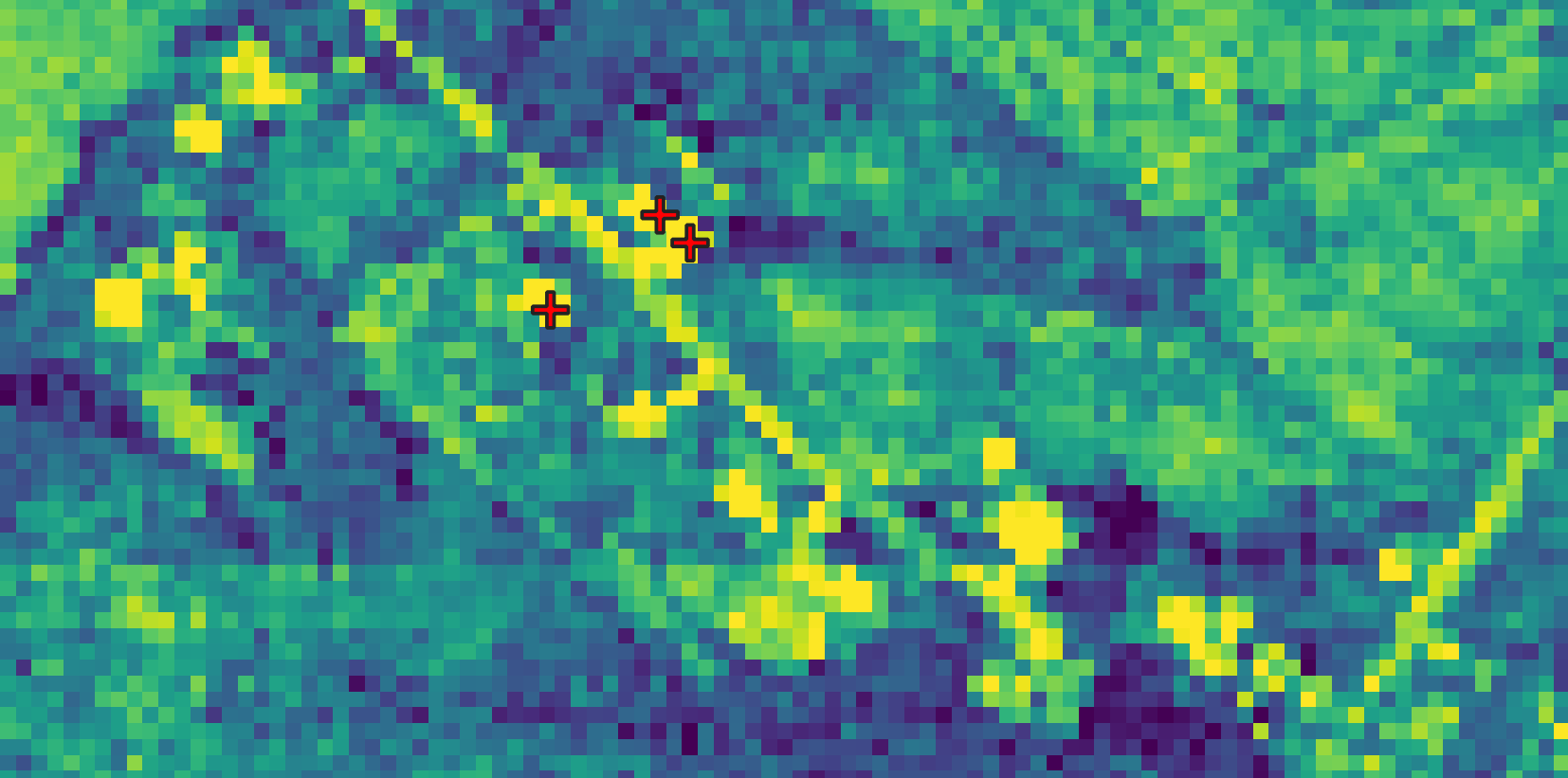} & &
     \includegraphics[width=0.45\linewidth]{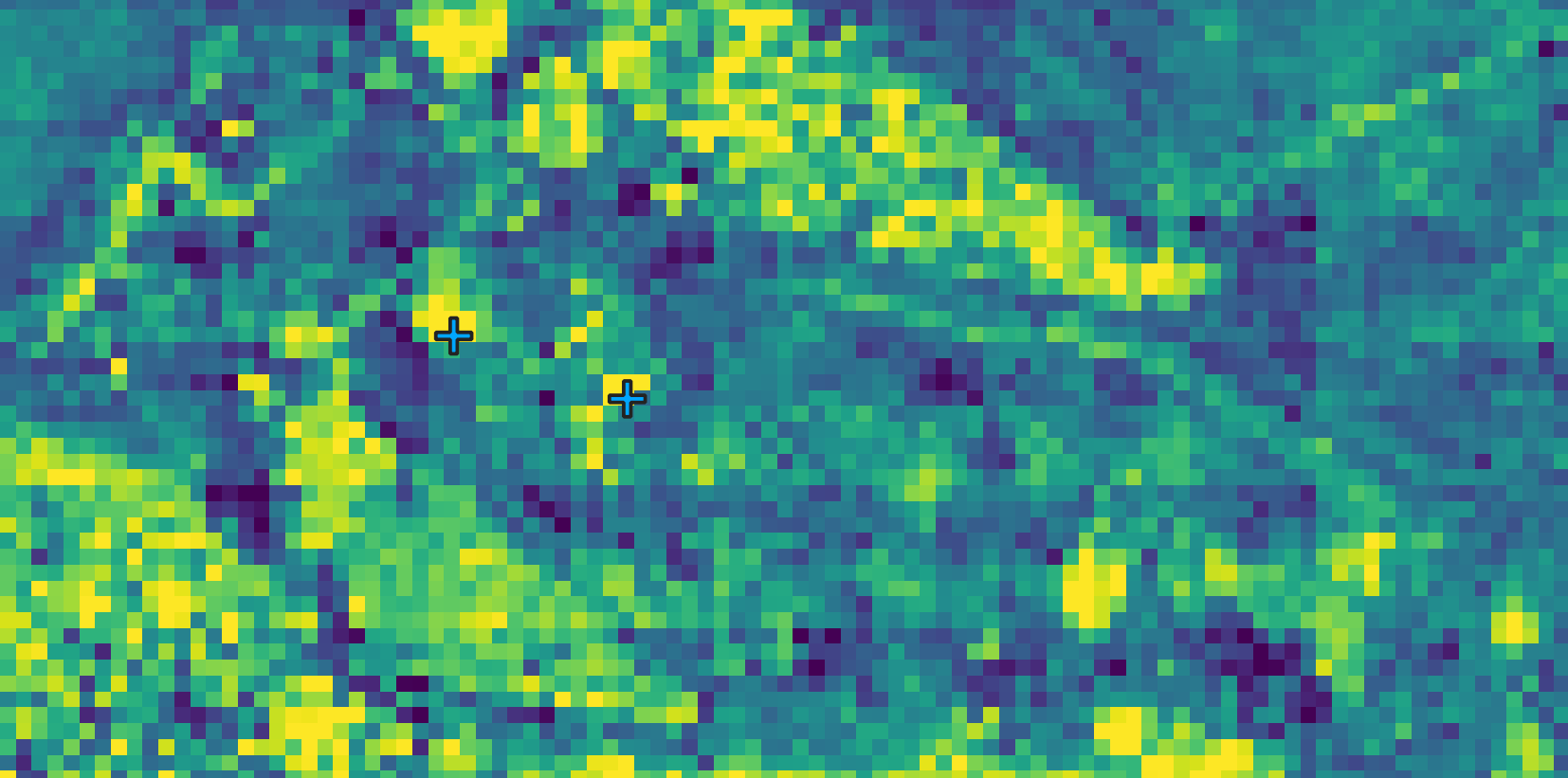}\\
\end{tabular}
    \caption{Example usage of cosine similarity with \href{https://catalog.data.gov/dataset/national-agriculture-imagery-program-naip}{NAIP data}. Using only a couple of points, we can identify trees or houses without needing to train a model for this specific task. From left to right and top to bottom: Original RGB data and provided template points (red and blue crosses); Features produced by a ViT DINO small encoder \citep{caron2021emerging}; Heatmap produced with the red points as input (houses); Heatmap produced with the blue points as input (trees with red leaves).}
    \label{fig:sim}
\end{figure}

When exploring high-dimensional spaces, similarity search is a common task. The similarity approach module of IAMAP enables users to generate similarity maps based on one or more point shapefiles. This module relies on cosine similarity, which assigns a score between 0 and 1 to two points based on their coordinates in the feature space. The score is zero if the vectors represented by these coordinates are orthogonal to the reference vectors provided by the user, and 1 if they are identical. This approach is commonly used for instance retrieval tasks in deep learning \citep{chen2022deep}, as it helps identify points that are closely represented in the feature space (see Fig.~\ref{fig:sim} for examples). By applying a threshold, this method can also be used for simple segmentation tasks.

\subsection{Supervised machine learning}

The final module of IAMAP enables users to build supervised predictive models using classical machine learning algorithms such as Random Forests, KNN, or Gradient Boosting. In contrast to other plugins that enable the use of end-to-end deep learning models for specific tasks \citep[\textit{e.g.}][]{aszkowski2023deepness}, we have focused on lighter machine learning algorithms to minimize dataset and computational resource requirements.
These algorithms often require relevant input features to be able to perform. While deep learning is now more potent on a lot of tasks, ML algorithms used with deep learning features as input can achieve satisfactory performances with a fraction of the cost needed to fit the algorithm. Then for example, algorithms such as KNN are used in deep learning research to evaluate models trained in an unsupervised way without having to retrain an entire model (\textit{e.g.} see SM of \cite{caron2021emerging}).

The plugin provides a wide array of available algorithms, once again using the \textit{scikit-learn} library as a back-end. More specifically, all methods provided by the \textit{ensemble} and \textit{neighbors} modules that share a common API are available.

Because this module relies on supervised approaches, it requires the user to provide a reference point dataset as a shapefile. The sampling design of this reference dataset is left to the user but we provide the option to choose how the validation scheme is performed. By default, a cross validation in k-fold is performed by randomly splitting the dataset into 5 folds. Otherwise, it is possible to define the train/test split or the cross-validation scheme dataset according to the values in a attribute column. As the appropriate validation scheme depends largely on the dataset and target task, this validation scheme might not be the most appropriate (see \cite{ploton2020spatial} for discussion on this topic in the context of spatial datasets). We therefore encourage the users to consider their choices of validation scheme via the plugin interface. 

\section{Usage example}
\label{sec:examples}

\begin{figure}
    \centering
    \includegraphics[width=1\linewidth]{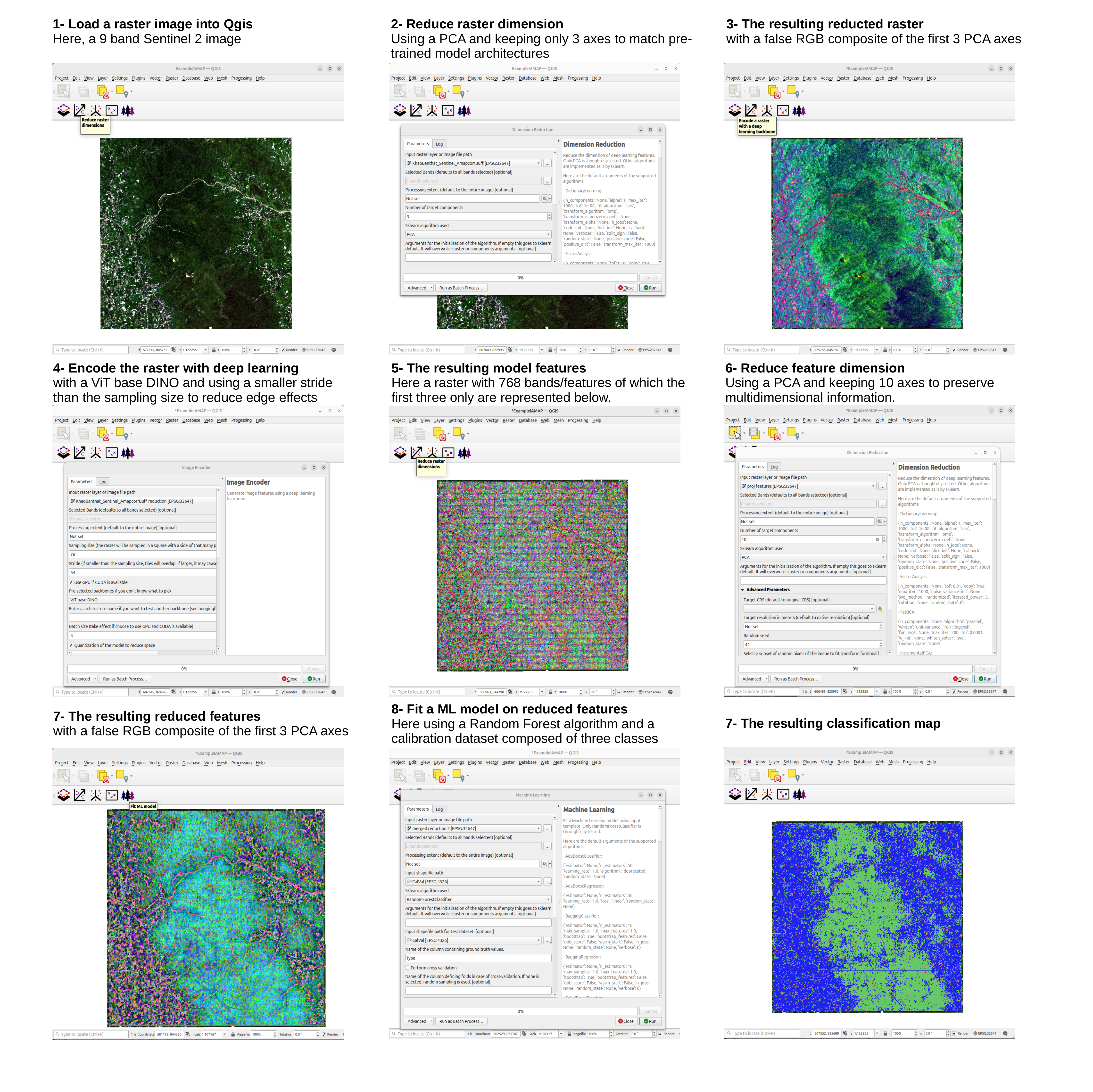}
    \caption{An example of workflow implemented using IAMAP to produce a ca. 50-m classification map from a 10-m multispectral Sentinel 2 image over a forested landscape from Thailand (Lat 7.53°, Lon 99.82°).}
    \label{fig:exampleIamap}
\end{figure}

Because the IAMAP plugin consists of a set of different modules that can be implemented independently or sequentially in various combinations, the number of possible uses is very large. Here, we provide one example of a potential workflow to produce a classification map using three complementary IAMAP modules (Fig. ~\ref{fig:exampleIamap}). Several other use cases, along with detailed protocols, are available \href{https://iamap.readthedocs.io/en/latest/examples.html}{in the online documentation of the plugin}.

\section{Design choices}

We have aimed the development for the plugin to be usable on a laptop without a GPU by someone with no coding experience. This has come with various design choices.

\subsection{Cross-platform and easy to install}

The plugin is designed to be easy to install, especially working with state of the art deep learning dependencies. Dependencies are handled using \textit{pip} with a startup script that automatically looks for dependencies and installs the missing one if needed. If needed, a \textit{conda} environment formula is provided as well to work in a separate fixed environment.
The plugin has been tested on Windows, Mac and Linux with several QGIS versions.
Although the plugin does not require a GPU to function, if one is detected, the correct version of \textit{pytorch} is downloaded to be used during deep learning inference. The user may opt out the usage of the GPU afterwards.

\subsection{Inference as a stoppable background task}

The inference of deep learning models on large raster images may be long, in particular without a GPU. Then, we have given the option to schedule small pauses during the inference, which limits the CPU usage and enables to use the PC for other tasks during the inference. 
An other choice has been to save batches on disk rather than keeping all inferred tiles on RAM. While slower, this makes possible to stop the inference and start again latter (even after reboot). Temporary files are cleaned up after use.

The produced rasters can become heavy and are therefore compressed by default to save space. 

\subsection{Model quantization}

The quantization of a deep learning model is the act of switching the encoding of the weights from \textit{float32} to a lighter format such as \textit{uint8}. This greatly reduces model size and inference time, at the cost of some precision (see \href{https://pytorch.org/blog/introduction-to-quantization-on-pytorch/}{Pytorch documentation}) \citep{wu2020integer}. Recently, DeepSeek AI have been able to divide training costs by 40 by relying on similar methods with \textit{fp8} precision training \citep{liu2024deepseek}.
This practice is common when working with hardware size constraints. Here, we give the option to the user to quantize the model before inference. When working with a model that was not specifically trained for the task asked of it, the trade-off between speed and precision could be beneficial more often than not. 


\section{Perspectives and future developments}

\subsection{Limitations of the plugin}
\label{sec:limitations}

This plugin is though for a usage in conditions where the end-to-end training of a neural network is not a possibility because of a lack of data or computing power. This comes with limitations to what is possible with deep learning in inference only compared to what can be achieved with neural network trained classically.

First, some task will require non-linear and complex connections in the feature space and will not be possible with simple manipulations as those possible with this plugin. For example, complex tasks as instance segmentation is easily achievable with dedicated deep learning models (see \cite{zhao2023geosam}) but not with our plugin.

By using deep learning methods, classical machine learning and data manipulation methods, this plugin inherits from advantages but also drawbacks from different types of algorithms. Ideally and depending on the  use case, the use of a deep learning encoder will provide relevant features, robust to low level noise and transformations. These features can then be leveraged with lighter machine learning algorithms, enabling the creation of maps that would not be possible without the features provided by a deep learning encoder. On the other hand, it may be required to test a variety of encoders and hyper-parameters to achieve satisfying results. While projection or clustering techniques are often easy to fit, testing different deep learning models can be time consuming, especially on restricted hardware.

\subsection{Future developments}

Future developments for the plugin include keeping up with computer vision state of the art but also optimization techniques to ensure lightweight inference time and usability on restricted hardware.

Moreover, we aim to implement more models dedicated to remote sensing tasks (for instance, those evaluated by \cite{marsocci2024pangaea}).
As of now, the feature extraction tool is though for ViT like encoders, that have spatially explicit features. We aim to develop it to be more generalist and take any encoder as input, such as ResNets or UNets that are still widely used in deep learning and remote sensing.

\section{Availability}

Development of the plugin is open sourced on GitHub \url{https://github.com/umr-amap/iamap}. Documentation is available at \url{https://iamap.readthedocs.io/}. The plugin is developed in continuous integration. We plan to publish the plugin on official QGIS repository to further ease the installation process. 

\section*{Acknowledgments}

The authors would like to thank all people who have tested this software during development and have provided meaningful feedback.

\section*{Conflict of interest}

The authors declare no conflict of interest.

\bibliographystyle{unsrtnat}
\bibliography{references}

\end{document}